\documentclass[10pt, twocolumn]{article}

\usepackage[utf8]{inputenc}
\usepackage[T1]{fontenc}
\usepackage{geometry}
\usepackage{mathpazo} 
\usepackage{amsmath, amssymb, amsfonts}
\usepackage{microtype} 
\usepackage{bm} 

\usepackage{titlesec}
\usepackage{abstract}
\usepackage{authblk}
\usepackage{hyperref}
\usepackage[font=small,labelfont=bf]{caption}

\titleformat{\section}{\large\bfseries}{}{0pt}{}
\titleformat{\subsection}{\normalsize\bfseries}{}{0pt}{}
\titleformat{\subsubsection}{\normalsize\itshape}{}{0pt}{}

\title{\Large The Geometry of Benchmarks: A New Path Toward AGI}

\author{Przemyslaw Chojecki}
\affil{ulam.ai}
\date{December 3, 2025}

\begin{document}

\twocolumn[
  \begin{@twocolumnfalse}
    \maketitle
    \begin{abstract}
    Benchmarks are the primary tool for assessing progress in artificial intelligence (AI), yet current practice evaluates models on isolated test suites and provides little guidance for reasoning about generality or autonomous self-improvement. Here we introduce a geometric framework in which \textit{all} psychometric batteries for AI agents are treated as points in a structured moduli space, and agent performance is described by capability functionals over this space. First, we define an \textbf{Autonomous AI (AAI) Scale}, a Kardashev-style hierarchy of autonomy grounded in measurable performance on batteries spanning families of tasks (for example reasoning, planning, tool use and long-horizon control). Second, we construct a \textbf{moduli space of batteries}, identifying equivalence classes of benchmarks that are indistinguishable at the level of agent orderings and capability inferences. This geometry yields determinacy results: dense families of batteries suffice to certify performance on entire regions of task space. Third, we introduce a general \textbf{Generator-Verifier-Updater (GVU)} operator that subsumes reinforcement learning, self-play, debate and verifier-based fine-tuning as special cases, and we define a self-improvement coefficient $\kappa$ as the Lie derivative of a capability functional along the induced flow. A variance inequality on the combined noise of generation and verification provides sufficient conditions for $\kappa > 0$. Our results suggest that progress toward artificial general intelligence (AGI) is best understood as a flow on moduli of benchmarks, driven by GVU dynamics rather than by scores on individual leaderboards.
    \vspace{0.5cm}
    \end{abstract}
  \end{@twocolumnfalse}
]

\section*{Introduction}

Modern AI systems are usually evaluated by their performance on fixed benchmarks - standardized test suites for language modelling, reasoning, vision, or control \cite{benchmarks1}. Performance is reported as a percentage score or leaderboard rank on individual datasets, occasionally aggregated across a small collection of tasks. This practice has three limitations.

First, benchmarks are typically \textit{narrow}: a system may excel on a particular dataset while failing catastrophically on nearby tasks \cite{narrowAI}. Second, evaluation is often \textit{fragmented}: each new domain introduces bespoke test suites, with little understanding of how different benchmarks relate to one another or how much new information they provide. Third, benchmarks are usually \textit{static}: they measure capabilities at a single point in time, while progress in AI-especially toward artificial general intelligence (AGI)-is fundamentally about \textbf{self-improvement over time} \cite{sutton}.

Here we develop a framework that addresses all three issues simultaneously by treating benchmarks themselves as mathematical objects and studying their \textbf{geometry}. The central idea is that once we consider \textit{all} psychometric batteries at once, they organize into a well-structured \textbf{moduli space}. Agents are then characterized not only by their scores on individual benchmarks but by their position and trajectory in this space.

Our contributions are threefold:
\begin{enumerate}
    \item We propose an \textbf{Autonomous AI (AAI) Scale}, an operational, Kardashev-inspired hierarchy that measures the autonomy and generality of AI systems. The AAI scale is defined in terms of performance on families of batteries under explicit resource constraints, providing a behavioural notion of ``AGI-like'' capacity. \cite{AAIscore}
    \item We develop a \textbf{moduli-theoretic view of batteries}. By quotienting out natural equivalence relations between test suites, we define a moduli space on which capability functionals become smooth fields, enabling geometric reasoning about generalization, coverage and redundancy of benchmarks. \cite{psychometric}
    \item We extend this static geometry to \textbf{dynamics of self-improvement}. We introduce a general \textbf{Generator-Verifier-Updater (GVU)} loop that subsumes reinforcement learning (RL), self-play, debate, adversarial training and verifier-based fine-tuning. Viewing GVU as a stochastic flow on the parameter manifold of agents, we define a self-improvement coefficient $\kappa$ and derive a \textbf{variance inequality} giving sufficient conditions for $\kappa > 0$ in terms of the noise in generation and verification. \cite{gvu}
\end{enumerate}

Together, these ingredients recast progress toward \textbf{AGI as a flow on moduli of batteries}. ``Everything is reinforcement learning'' in the sense that any practical self-improving AI system can be represented as a GVU flow that climbs a capability functional defined over this moduli space.

\section*{Results}

\subsection*{1. An Autonomous AI Scale}

\subsubsection*{1.1 Agents, tasks and batteries}
We formalize an \textit{agent} as a policy $\pi$ acting in interactive environments, mapping histories of observations and actions to distributions over actions. A \textbf{task instance} $\tau$ specifies an environment, initial condition and termination rule, together with a scoring functional
\begin{equation}
    s_\tau(\pi) \in [0,1]
\end{equation}
that measures the performance of $\pi$ on $\tau$ given a resource budget (for example, number of calls, computation time or human interventions).

A \textbf{battery} $\mathcal{B}$ is a finite or countable collection of task instances equipped with:
\begin{itemize}
    \item a sampling distribution $\mu_{\mathcal{B}}$ over tasks and random seeds,
    \item a scoring rule that aggregates instance-level scores into a \textbf{battery score}
    \begin{equation}
        S(\pi;\mathcal{B}) = \mathbb{E}_{\tau \sim \mu_{\mathcal{B}}} \left[ s_\tau(\pi) \right],
    \end{equation}
    \item metadata specifying the family of capabilities probed (e.g., mathematical reasoning, tool orchestration, long-horizon planning).
\end{itemize}

We write $F(\pi, \mathcal{B})$ for a \textbf{capability functional} which may coincide with $S$ or incorporate penalties for resource usage and uncertainty. 
Formally, such functionals are acting on the agent's trajectory distribution under $\mathcal{B}$ and are required to satisfy the axioms of naturality, restricted monotonicity, threshold calibration and generality.

\subsubsection*{1.2 A multi-axis AAI Index}
Real-world agency is multi-dimensional. We therefore consider a finite collection of \textit{families} $\mathcal{F} = \{f_1,\dots,f_m\}$ (for example reasoning, learning, memory, tool use, social interaction). For each family $f$ we specify a set of batteries $\mathcal{B}_f$ that probe that family under standardized protocols.

The \textbf{AAI Index} of an agent $\pi$ is the vector
\begin{equation}
    \text{AAI}(\pi) = \big( F_f(\pi) \big)_{f \in \mathcal{F}}, \quad F_f(\pi) = \inf_{\mathcal{B} \in \mathcal{B}_f} F(\pi, \mathcal{B}),
\end{equation}
which measures worst-case performance over batteries in each family, under fixed resources. When desired, a scalar AAI score can be obtained by aggregating with a monotone functional (for example a norm or weighted quantile), but the vector form is primary.

\subsubsection*{1.3 Level gates and a Kardashev-style hierarchy}
To make the scale actionable, we define discrete \textbf{AAI levels} by specifying \textbf{level gates}: for level $\ell$ and each family $f$, a threshold $\theta_{f,\ell}$ and robustness parameters (for example tolerated degradation under perturbations of the battery).

An agent $\pi$ is said to be at \textbf{level $\ell$} if, for all families $f$,
\begin{equation}
    F_f(\pi) \ge \theta_{f,\ell},
\end{equation}
and this inequality continues to hold under (controlled) drifts in battery composition and environment parameters. We denote the highest level satisfied by $\pi$ as AAI-$\ell$.

Instantiations of this scheme yield interpretable levels such as:
\begin{itemize}
    \item \textbf{AAI-0:} Narrow, non-autonomous systems that require human oversight on every step.
    \item \textbf{AAI-1:} Task-specific, tool-using systems with limited transfer across families.
    \item \textbf{AAI-2:} Broadly capable assistants with moderate autonomy but short-horizon memory.
    \item \textbf{AAI-3:} Open-world agents that can plan, learn and adapt across many families.
    \item \textbf{AAI-4 ("AGI"):} Systems exhibiting sustained, autonomous self-improvement ($\kappa > 0$; see below) across diverse batteries-a behavioural analogue of ``superintelligent'' performance.
\end{itemize}
In this way, AGI-like behaviour is operationalized as crossing a set of performance gates defined on batteries rather than as passing a single monolithic test.

\subsection*{2. Moduli of Psychometric Batteries}

\subsubsection*{2.1 Equivalence of batteries}
Different benchmarks often probe the same underlying capabilities. We capture this by an \textbf{equivalence relation} on batteries.

Two batteries $\mathcal{B}_1, \mathcal{B}_2$ are said to be \textbf{evaluation-equivalent} if they differ only by the obvious symmetries: permuting tasks within families, relabelling seeds and drifts in a measure-preserving way, applying strictly increasing reparameterizations to scores, and rescaling resource units by positive constants, in a way that preserves all evaluation statistics. 

The equivalence classes under this relation are \textbf{moduli of batteries}. We denote the set of classes by $\mathcal{M}$.

\subsubsection*{2.2 Geometry on the moduli space}
We endow $\mathcal{M}$ with additional structure:
\begin{itemize}
    \item A \textbf{topology} induced by the Wasserstein metric on canonical score laws.
    \item A \textbf{distance} $d([\mathcal{B}_1],[\mathcal{B}_2])$ given by that metric on canonical representatives.
    \item A decomposition into \textbf{regions} corresponding to families of capabilities.
\end{itemize}

On this space, a capability functional $F(\pi,\mathcal{B})$ induces a smooth field
\begin{equation}
    \Phi_\pi : \mathcal{M} \to \mathbb{R}, \quad \Phi_\pi([\mathcal{B}]) = F(\pi, \mathcal{B}),
\end{equation}
well-defined on equivalence classes. For a fixed $\pi$, the variation of $\Phi_\pi$ across $\mathcal{M}$ describes the agent's sensitivity to different kinds of evaluation; for fixed $[\mathcal{B}]$, the variation across a family $(\pi _t)_t$ describes the discriminative power of the battery.

\subsubsection*{2.3 Determinacy and coverage}
The geometric viewpoint yields \textbf{determinacy results}: under mild regularity assumptions, a dense subset of batteries suffices to approximate an agent's capability on a whole region of $\mathcal{M}$.

Informally, suppose that for a region $U \subset \mathcal{M}$ and a dense subset $\{[\mathcal{B}_i]\}_{i\in I} \subset U$, the fields $\Phi_\pi$ are Lipschitz in the metric $d$. Then, for any $[\mathcal{B}] \in U$,
\begin{equation}
    \big|\Phi_\pi([\mathcal{B}]) - \Phi_\pi([\mathcal{B}_i])\big| \le L \cdot d([\mathcal{B}],[\mathcal{B}_i]),
\end{equation}
for some constant $L$ and suitable $i$. Thus, sufficiently rich batteries in $U$ provide \textit{coverage} of the region; testing on all of them is unnecessary.

Operationally, this allows the design of benchmark suites that provide \textbf{maximal information} for a given evaluation budget, by selecting representatives that are far apart in $\mathcal{M}$ while remaining in the region of interest.

\subsection*{3. Self-Improvement as GVU Dynamics}

\subsubsection*{3.1 The Generator-Verifier-Updater loop}
Practical AI systems are rarely static. We abstract training, fine-tuning and adaptation into a general \textbf{Generator-Verifier-Updater (GVU)} loop. 

We thus consider a time-indexed family of agents $(\pi_t)_{t}$ representing successive versions of the same system during learning. At iteration $t$, we write $\theta _t \in \Theta$ for the current parameters on the parameter manifold $\Theta$ and $\pi_t := \Pi_{\Theta}(\theta_t)$ for the induced policy where $\Pi_{\Theta}$ is the architecture map. 
For such a family, a single update step $\pi_t \to \pi_{t+1}$ can (up to parametrization) be written as the composition of three maps:
\begin{enumerate}
    \item \textbf{Generator (G):} produce candidate trajectories, outputs or parameter proposals $z_t \sim G(\pi_t)$. Examples include rollouts in an environment, sampled responses to prompts, or candidate weight updates.
    \item \textbf{Verifier (V):} evaluate the candidates against a (possibly stochastic) signal $r_t = V(z_t)$. This encompasses rewards in RL, human feedback, automatic checkers, discriminators, theorem provers and other evaluators.
    \item \textbf{Updater (U):} update the agent using the evaluated candidates, yielding a new parameter state
    \[
      \theta_{t+1} = U(\theta_t, z_t, r_t),
    \]
    and hence a new policy $\pi_{t+1} := \Pi_{\Theta}(\theta_{t+1})$. Instances include gradient steps, policy iteration, rejection sampling, prompt editing or architectural changes.
\end{enumerate}

Composition of these steps defines a GVU operator
\[
  \mathcal{T}_{\mathrm{GVU}} : \Theta \to \Theta,
  \qquad
  \theta_{t+1} = \mathcal{T}_{\mathrm{GVU}}(\theta_t),
\]
which induces a discrete-time flow $(\theta_t)_{t}$ on the parameter manifold and an associated family of policies $(\pi_t)_{t}$ via the architecture map $\Pi_{\Theta}$. In the limit of small step sizes, this flow can be approximated by a stochastic differential equation on the statistical manifold $(\Theta,g)$, where $g$ is chosen to be the Fisher information metric induced by the policy family.
\subsubsection*{3.2 Capability functionals and the self-improvement coefficient}
Let $F(\pi,\mathcal{B})$ be a capability functional for a fixed equivalence class $[\mathcal{B}]$ of batteries. For a time-indexed family of agents $(\pi_t)_t$ we have corresponding capability trajectory $F_t = F(\pi_t, \mathcal{B})$.

Assuming sufficient regularity and viewing the discrete GVU updates as approximating a continuous-time flow $t \mapsto \pi_t$, we define the \textbf{self-improvement coefficient} $\kappa_t$ as the instantaneous rate of change
\begin{equation}
    \kappa_t := \frac{d}{dt} F(\pi_t, \mathcal{B}).
\end{equation}
Equivalently, $\kappa_t$ is the Lie derivative of $F$ along the drift field induced by the GVU dynamics on policy space. 

Positive $\kappa$ indicates that the agent is improving on the battery (and, by determinacy, on a region of moduli space); $\kappa \approx 0$ corresponds to a plateau; negative $\kappa$ indicates degradation (for example due to mode collapse or overfitting).

\subsubsection*{3.3 A variance inequality for GVU}
The central theoretical result of the GVU formalism is a \textbf{variance inequality} that provides sufficient conditions for $\kappa > 0$ in terms of alignment and noise.

In outline, suppose that:
\begin{itemize}
    \item The updater approximates a stochastic gradient ascent step on $F$ with learning rate $\eta$.
    \item Generator and verifier introduce zero-mean noise with covariance $\Sigma_G$ and $\Sigma_V$ in the gradient estimate.
    \item The Hessian of $F$ and the curvature of the parameter manifold satisfy mild boundedness assumptions.
\end{itemize}

Then the expected change in capability over one step can be decomposed as
\begin{multline}
    \mathbb{E}\big[F(\pi_{t+1},\mathcal{B}) - F(\pi_t,\mathcal{B})\big] \\
    \approx \eta \, |\nabla F(\pi_t,\mathcal{B})|^2 - \frac{\eta^2}{2} \, \mathrm{Tr}\big(H_F(\pi_t) \Sigma_{\text{GV}}\big) - O(\eta^3),
\end{multline}
where $\Sigma_{\text{GV}}$ is a positive semidefinite combination of $\Sigma_G$ and $\Sigma_V$. The first term drives improvement; the second term captures degradation due to noise and curvature.

The \textbf{variance inequality} asserts that for step sizes $\eta$ in a non-trivial interval $(0, \eta^*)$, the expected improvement is positive whenever
\begin{equation}
    \mathrm{Tr}\big(H_F(\pi_t) \Sigma_{\text{GV}}\big) < c \, |\nabla F(\pi_t,\mathcal{B})|^2,
\end{equation}
for a constant $c$ depending on curvature bounds. Intuitively, as long as the gradient signal dominates the combined noise of generator and verifier, there exists a regime of learning rates for which $\kappa > 0$.

\subsubsection*{3.4 ``Everything is reinforcement learning''}
Many seemingly distinct training schemes become special cases of GVU:
\begin{itemize}
    \item \textbf{Standard RL:} G samples trajectories, V computes returns, U performs policy updates.
    \item \textbf{Self-play:} G samples games between agent copies, V applies win/loss rules, U updates via policy gradients.
    \item \textbf{Language-model self-improvement:} G generates candidates, V scores via feedback/debate, U fine-tunes.
    \item \textbf{Adversarial training (GANs):} The discriminator plays the role of V.
\end{itemize}
In each case, learning can be seen as stochastic gradient ascent on a capability functional defined over some region of moduli space. From this perspective, \textbf{reinforcement learning is a geometric pattern of self-improvement}.

\section*{Discussion}

We have proposed a framework in which evaluation, generality and self-improvement of AI systems are described within a single geometric picture.

The \textbf{Autonomous AI Scale} \cite{AAIscore} replaces informal notions of ``AGI'' with performance thresholds on families of batteries. The \textbf{moduli space of batteries} \cite{psychometric} organizes benchmarks into equivalence classes, and the \textbf{GVU dynamics} \cite{gvu} connect this static geometry to learning, showing that many contemporary training procedures are special cases of reinforcement learning on the moduli space, with a variance inequality providing conditions for positive self-improvement.

Several directions follow. First, the formalism suggests \textbf{new ways to design benchmark suites}. Rather than collecting tasks ad hoc, one can aim for representatives that are far apart in moduli space, maximizing information per evaluation. Second, monitoring the \textbf{self-improvement coefficient $\kappa$} and the variance of verifiers could become part of safety protocols for autonomous agents \cite{safety}. Third, empirical work is needed to construct \textbf{practical embeddings of real-world benchmarks} into approximate moduli spaces, using techniques from representation learning and manifold modelling.

Ultimately, the geometric view implies that progress toward AGI is unlikely to be captured by performance on a single canonical test. Instead, it is a matter of how agents move through - and reshape - the space of tasks we use to probe them.

\section*{Methods (summary)}
\footnotesize

\textbf{1. Formal model of batteries and moduli.}
We model batteries as tuples
$\mathcal{B} = (T,\mathcal{F},\mathsf{S},Q^*,\mu,\mathsf{D},\Pi,\mathsf{R})$,
where $T$ is a finite set of tasks, $\mathcal{F}$ is a partition of $T$ into
families, $\mathsf{S}=\{S_t\}_{t\in T}$ are task-specific scoring maps,
$Q^*$ are task thresholds, $\mu$ is a sampling law on $T\times\Pi\times\mathsf{D}$
(tasks, seeds, drifts), $\mathsf{D}$ is a space of drifts, $\Pi$ is a space of
seeds, and $\mathsf{R}$ are resource coordinates (for example time or cost).
An agent and battery together induce a canonical evaluation law on an
evaluation space $X_{\mathcal{B}}$ and a capability functional
$F(\pi,\mathcal{B}) = \Phi_{\mathcal{B}}(\rho_{\mathcal{B}}(\pi))$ as in
\cite{psychometric}. Moduli are obtained by quotienting the space of such
batteries by the evaluation-equivalence relation defined in the main text
(measure-preserving relabellings, monotone score reparameterizations, and
resource rescalings that leave all capability comparisons invariant).

\textbf{2. Topology and metric.}
The moduli space $\mathcal{M}$ is endowed with the topology and distance induced by the Wasserstein metric on the canonical score laws.

\textbf{3. Determinacy theorems.}
Under Lipschitz regularity of $F$, we show that finite $\varepsilon$-nets in $\mathcal{M}$ suffice to approximate capability fields uniformly. Proofs use standard covering arguments and concentration inequalities for empirical estimates of battery scores.

\textbf{4. Information-geometric flows.}
Agent parameters $\theta$ live on a Riemannian manifold $(\Theta,g)$ with metric given by the Fisher information. GVU updates are modelled as stochastic differential equations
$d\theta_t = v(\theta_t)dt + \sigma(\theta_t)dW_t$.
The self-improvement coefficient $\kappa$ is expressed as the Lie derivative of $F$ along $v$, and the variance inequality can be 
understood as arising from an Itô-style expansion of $F$ along this 
flow, together with curvature bounds on $(\Theta,g)$ and spectral 
estimates on $\sigma\sigma^\top$.

\textbf{5. Data and code availability.} No datasets were generated or analysed in this study.

\bibliographystyle{naturemag}

\end{document}